\documentclass[letterpaper,twocolumn,10pt]{article}
\usepackage{usenix}

\usepackage{tikz}
\usepackage{amsmath}

\usepackage{filecontents}
\usepackage{tikz}
\usepackage{amsmath}
\usepackage{enumitem}
\usepackage{multirow}
\usepackage{pifont}
\usepackage[utf8]{inputenc}
\usepackage[russian,english]{babel}
\usepackage{textgreek}
\usepackage{subcaption}
\usepackage{hyperref}
\usepackage{tcolorbox}
\usepackage{amssymb}
\usepackage{graphicx}
\usepackage{pifont}

\begin{document}

\date{}

\title{Detect, Unlearn, Restore: Defending Text Summarization Models \\ Against Data Poisoning}

\author{
{\rm Poojitha Thota}\\
{\rm The University of Texas at Arlington}\\
{\rm Arlington, Texas, USA}\\
\texttt{\rm poojitha.thota@mavs.uta.edu}
\and
{\rm Shirin Nilizadeh}\\
{\rm The University of Texas at Arlington}\\
{\rm Arlington, Texas, USA}\\
\texttt{\rm shirin.nilizadeh@uta.edu}
} 

\maketitle

\begin{abstract}
Training-time data poisoning during fine-tuning poses a significant threat to large language models (LLMs) deployed for abstractive text summarization, where small task-specific datasets exert disproportionate influence on model behavior. 
In this setting, adversaries manipulate fine-tuning data to induce persistent summarization failures, such as biased or harmful summaries, while preserving standard evaluation metrics.
We present a unified post-hoc defense framework for detecting and remediating fine-tuning-stage poisoning in summarization models across the machine learning supply chain. 
Our experiments show that in white-box settings, poisoned document–summary pairs exhibit abnormally high training influence, enabling detection via influence-function analysis with semantic consistency checks. 
In black-box settings, poisoned models display two-to-three times greater sensitivity to semantics-preserving perturbations, enabling behavioral auditing without training data access.
Beyond existing poisoning formulations, we introduce novel attacks targeting factual distortion and representational bias, showing that poisoning alters summarization behavior without triggering conventional alarms. 
Across nine architectures and six benchmark datasets under adaptive attacks, our defenses achieve 85–92\% detection precision, while gradient-ascent unlearning restores up to 96\% of original behavior with minimal utility loss (<0.6\% ROUGE degradation). 
These results indicate that fine-tuning-time poisoning leaves persistent structural artifacts, enabling practical detection and post-deployment recovery without full retraining.
\end{abstract}

\section{Introduction}
The widespread adoption of Large Language Models (LLMs) has revolutionized natural language processing (NLP), with text summarization emerging as one of the most impactful applications for efficient information processing and knowledge management~\cite{ravaut2024context, dagdelen2024structured, liu2024learning}. Recent advances in LLMs such as GPT-4~\cite{hurst2024gpt}, Claude~\cite{anthropic_claude_4}, and PaLM~\cite{anil2023palm} have demonstrated remarkable capabilities in generating coherent, context-aware summaries from vast textual data. From news aggregation platforms to scientific literature reviews, abstractive text summarization models have become integral to how individuals and organizations process large volumes of information at scale~\cite{zhang2022comprehensive}. 
As these models become integral to critical decision-making systems, recent work has shown pressing security concerns, including fundamental weaknesses in summarization models and their susceptibility to data-poisoning attacks~\cite{ebrahimi2017hotflip, li2023efficiently, wang2022semattack, thota2024attacks}.

In this work, we focus specifically on poisoning that occurs during the fine-tuning stage, where domain-specific datasets are small, and each example exerts disproportionately high gradient influence. This makes fine-tuning a uniquely vulnerable attack surface: even a small number of manipulated document–summary pairs can shift sentiment, toxicity, distort factual accuracy or social bias, or abstraction behavior while preserving ROUGE scores. Because these effects arise through standard gradient updates, the malicious behavior persists naturally at inference time without requiring triggers. 

We consider two attack vectors: (1) adversaries release poisoned datasets that victims use for fine-tuning (white-box), and (2) adversaries directly publish poisoned model checkpoints that appear normal on benchmarks but exhibit manipulated behavior in real use (black-box). These two settings impose different constraints on defenders. 
Within these vectors, we evaluate four poisoning objectives. 
Following prior work~\cite{thota2024attacks}, we include sentiment inversion and toxic injection attacks, and we additionally introduce two more attacks, \emph{factual distortion} and \emph{representational bias}, that corrupt summary correctness and demographic fairness without relying on explicit sentiment or toxicity cues. Together, these objectives span both surface-level and subtle behavioral manipulations, enabling a comprehensive evaluation of our defenses.

Existing defenses against adversarial attacks~\cite{goodfellow2014explaining,madry2018towards,moraffah2024adversarial}, text classification poisoning~\cite{shentextshield,liu2020joint,peitextguard}, and backdoor detection~\cite{wang2019neural,gao2019strip} do not address training-time poisoning in generative summarization, which involves trigger-free behavioral shifts rather than perturbation robustness or discrete misclassification. Recent defenses for harmful fine-tuning~\cite{sun2025peftguard,zhao2024defending,chen2025sdd,huangantidote,chapagain2025pruning} target safety alignment or classification tasks. While machine unlearning~\cite{bourtoule2021machine} and influence functions~\cite{koh2017understanding,kwondatainf} enable selective sample removal, they have not been systematically applied to defend against data poisoning in text summarization.

To address these challenges, we propose a unified defense framework with two complementary mechanisms, targeting distinct threat scenarios in the fine-tuning pipeline (Section~\ref{sec:threat}). 
Against white-box attacks where adversaries poison training datasets, our \textbf{Defense-1 (Poisoned Dataset Detection)} combines influence function analysis using DataInf~\cite{kwondatainf} with a modular behavioral filtering stage that detects abnormal semantic signals (e.g., sentiment shifts, toxicity, factual inconsistencies, or representational bias), followed by gradient-ascent unlearning~\cite{yao2024machine} to selectively remove poisoned samples. 
Our key insight is that effective poisoned samples tend to have a high influence, enabling detection by focusing on a small, high-impact subset rather than inspecting entire datasets. 
Against black-box attacks where adversaries distribute poisoned 
models, our \textbf{Defense-2 (Poisoned Model Detection)} exploits a novel compound vulnerability: poisoned models exhibit \textit{2--3$\times$ greater sensitivity} to adversarial perturbations than clean models.
Our Sensitivity to Adversarial Perturbations (SAP) metric, inspired by prior sensitivity-based detection in vision~\cite{fares2024attack}, quantifies lead-sentence exclusion under controlled perturbations, enabling detection without training data access.
Together, these two components operate as a single end-to-end framework, jointly protecting both poisoned datasets and distributed model checkpoints across the model lifecycle.

We conduct comprehensive evaluation across six datasets (MultiNews~\cite{fabbri2019multi}, CNN/DailyMail~\cite{chen2016thorough}, WikiSum~\cite{cohen2021wikisum}, ArXiv-summarization~\cite{cohan2018discourse}, PubMed-summarization~\cite{cohan2018discourse}, Multi-XScience~\cite{lu2020multi}) and nine models (encoder-decoder: BART-Large~\cite{lewis2019bart}, T5-Small~\cite{raffel2020exploring}, Pegasus-Large~\cite{zhang2020pegasus}, and FLAN-T5-Large~\cite{chung2024scaling}; open-source decoder-only models: LLaMA-3-8B~\cite{dubey2024llama}, Qwen-2.5~\cite{qwen2024qwen2}, Qwen-3~\cite{yang2025qwen3}, Mistral-7B~\cite{jiang2023mistral}, and Vicuna-13B~\cite{vicuna2023}). 
Defense-1 achieves 85.3\% average recovery while reducing remediation cost by 4–6$\times$ compared to full retraining, and consistently outperforms adversarial training, SISA~\cite{bourtoule2021machine}, exact retraining~\cite{guo2020certified}, selective pruning~\cite{liu-etal-2025-modality}, and TracIn~\cite{pruthi2020estimating}.
Defense-2 achieves near-perfect detection, in a fully black-box setting, outperforming CLIBE~\cite{zengclibe} and perplexity-score~\cite{rosenfeld1996maximum} based sensitivity baselines by maintaining clear separation between clean and poisoned models across all perturbation types.
Additionally, prior work~\cite{thota2024attacks} has shown that model poisoning can alter the fundamental behavior of summarization models, shifting them from abstractive to extractive generation. Critically, our defense \emph{reverses these behavioral shifts}. 
Beyond surface metrics, our defense restores deeper generation behavior: gradient-ascent unlearning recovers 94–96\% of abstractive behavior previously shifted toward extractiveness~\cite{thota2024attacks}, demonstrating that high ROUGE alone is insufficient to verify successful remediation.
This provides the first evidence that machine unlearning in Defense-1 restores deeper generation behavior. 

To validate robustness against defense-aware adversaries, we evaluate four adaptive attack strategies: low-influence poisoning, crafting samples with moderate influence scores, multi-objective poisoning, blending sentiment and toxicity, gradient masking, obscuring detection signals, and distributed poisoning, spreading attacks across batches. 
Defense-1 maintains 78–82\% recovery (vs. 85.3\% under standard attacks), while Defense-2 retains 95–98\% TPR. Statistical analysis confirms significant separation between clean and poisoned behaviors ($p < 0.001$), and results generalize across architectures.
The primary contributions of our work are: 
\ding{172} \textbf{First comprehensive defense framework} for text summarization against data poisoning using machine unlearning to target root causes, addressing both poisoned datasets and poisoned models scenarios.
\ding{173} \textbf{Defense-1: A novel influence-based poisoned sample detection and unlearning} method that identifies high-impact poisoned fine-tuning samples and selectively removes their influence without full retraining, enabling post-deployment remediation.
\ding{174} \textbf{Defense-2: A new poisoned model auditing via adversarial sensitivity (SAP metric)}, to detect poisoning-induced behaviors without access to training data.
\ding{175} \textbf{Novel attack formulations extending beyond existing poisoning objectives}, including factual distortion attacks and representational bias attacks for summarization, alongside sentiment and toxicity attacks, providing a comprehensive threat evaluation.
\ding{176} \textbf{First demonstration of behavioral recovery via unlearning}, achieving 94--96\% restoration of abstractive generation, establishing behavioral validation beyond output quality metrics.
\ding{177} \textbf{Extensive empirical validation} demonstrating Defense-1 achieves 85.3\% behavioral recovery (2.5--50\% contamination) with minimal utility degradation (ROUGE-1 $<$0.006), Defense-2 achieves perfect detection (100\% TPR, 0\% FPR), and both defenses maintain 78--82\% effectiveness against adaptive attacks, evaluated across 6 datasets, 9 models, and validated against 4 unlearning and 3 backdoor detection baselines.

\section{Related Work}
\label{sec:related_work}
Text-based models face two primary security threats: adversarial perturbations at inference time and data poisoning during training. 

\textbf{Inference-time Robustness.}
Adversarial training~\cite{goodfellow2014explaining,madry2018towards}, input sanitization~\cite{moraffah2024adversarial}, and NLP-specific extensions such as continuous relaxations~\cite{miyato2017adversarial}, curriculum-based perturbation schedules~\cite{wang2019improving}, and task-specific adaptations~\cite{jin2020bert}, improve robustness against perturbation-based attacks but do not address training-time manipulations that alter the model’s summarization behavior. 
Detection-based defenses based on perplexity shifts~\cite{rosenfeld1996maximum}, text purification~\cite{li2023text}, or randomized smoothing~\cite{cohen2019certified} similarly target perturbed inputs rather than corrupted training data. Backdoor detection approaches~\cite{wang2019neural,gao2019strip} further assume trigger-based misclassification, which does not capture poisoning-induced behavioral drift in generative summarization.

\textbf{Training-time Poisoning.}
Training-time data poisoning defenses have been extensively studied for text-based models. Existing approaches identify poisoned samples through spectral analysis~\cite{tran2018spectral}, activation clustering~\cite{chen2018detecting}, certified defenses~\cite{xu2021mitigating,dwork2014algorithmic}, or mitigate poisoning effects during training via adversarial training~\cite{miyato2017adversarial,wang2019improving,madry2018towards} or purification techniques~\cite{li2023text,moraffah2024adversarial}. 
Backdoor detection methods identify trigger-based attacks in text classification and NLU tasks~\cite{wang2019neural,gao2019strip,zengclibe}. 
\textit{\textbf{Machine unlearning}} provides another purification mechanism, removing specific training samples' influence without full retraining through gradient-based updates~\cite{thudi2022unrolling,golatkar2020eternal}, partition-based methods~\cite{bourtoule2021machine}, or influence functions~\cite{koh2017understanding,kwondatainf}. These defenses primarily target classification tasks with discrete labels and bounded outputs. 
Defenses for fine-tuned models present additional challenges. Recent work addresses post-fine-tuning safety alignment~\cite{huangantidote}, parameter-efficient backdoor detection and pruning~\cite{sun2025peftguard,zhao2024defending,chapagain2025pruning,liu-etal-2025-modality}, and training-time regularization~\cite{chen2025sdd,huadaptive,rosati2024representation,rosati2024immunization}. However, these approaches primarily target safety alignment failures in instruction-following or trigger-based backdoors requiring explicit patterns. In fine-tuned generative models like abstractive summarization, poisoning manifests as subtle semantic distortions during content selection and generation without discrete labels or explicit triggers, making existing detection and filtering strategies difficult to apply directly.
While prior unlearning work focuses on privacy compliance~\cite{cao2015towards,gupta2021adaptive,chourasia2023forget,nguyen2025survey,jang2023knowledge,yao2024machine}, our work is the first to systematically apply these methods to address security vulnerabilities in text summarization models, specifically targeting lead bias and data poisoning attacks.

\section{Threat Model}
\label{sec:threat}
We consider training-time data poisoning attacks against text summarization models, where an adversary inserts malicious or manipulated document–summary pairs into the fine-tuning corpus. Unlike inference-time adversarial attacks that affect individual inputs, poisoning during fine-tuning alters global model behavior and persists after deployment. 
The attacker aims to induce systematic shifts in generated summaries, while maintaining normal performance on standard metrics (e.g., ROUGE) so that the attack remains undetected during validation. 

\begin{figure}[t]
    \centering
    \includegraphics[width=0.8\columnwidth]{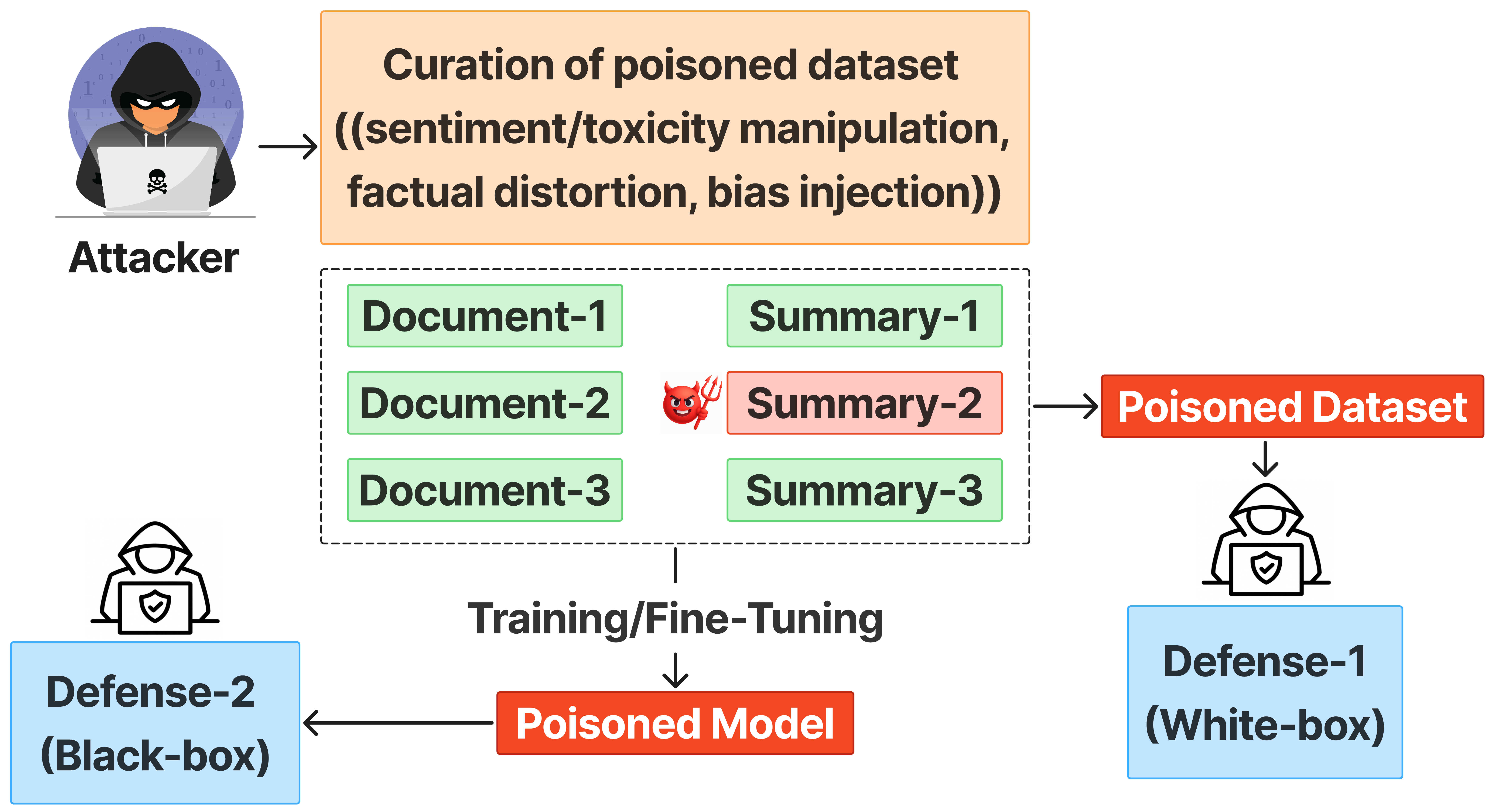}
    \caption{Threat model for training-time data poisoning in summarization. An attacker curates a poisoned dataset that either (i) is directly inspected by the defender (Defense-1, white-box) or (ii) is used to train and release a poisoned model (Defense-2, black-box).}
\label{fig:threat_model}
    \vspace{-0.7em}
\end{figure}

The attacker is assumed to construct poisoned samples that are linguistically fluent and distributionally consistent with the clean fine-tuning data, making manual inspection or simple filtering ineffective. The attacker cannot modify the victim’s training pipeline but relies on the poisoned data being incorporated during standard supervised fine-tuning. Poisoning does not rely on inference-time trigger phrases; instead, malicious behavior emerges naturally during summarization, distinguishing our threat model from backdoor attacks that require explicit triggers~\cite{wang2019neural}.
Poisoning yields two distinct defender environments, depending on which artifacts the attacker releases and what the defender can observe. As visualized in Figure~\ref{fig:threat_model}, the attacker-curated poisoned dataset can either be directly provided to the defender (Scenario 1) or used to train a poisoned model that is later released (Scenario 2).

\textbf{Scenario 1: White-Box (Poisoned Dataset):} The attacker releases a poisoned fine-tuning dataset. The defender has full access to this dataset and to the model architecture and training procedure. Although fine-tuning datasets are smaller than pretraining corpora, only a small fraction is poisoned (2.5\%--50\% in our evaluation), and poisoned samples are crafted to appear benign, making manual inspection or naive filtering unreliable. The defender must identify which examples are malicious without ground-truth annotations. In this white-box setting, defenders can compute gradients and influence scores, and perform sample-level attribution and targeted unlearning. 

\textbf{Scenario 2: Black-Box Inference-Only Setting (Poisoned Model):} 
In this scenario, the attacker releases a model that was trained on a poisoned dataset. The defender receives only the trained model checkpoint and has no access to the training corpus, training history, or data provenance. 
Although the defender can load and run the model locally, they still lack the information required to perform sample-level detection or retraining. The defender must therefore determine whether the model has been poisoned solely from its observable behavior during inference. 
Defense must rely exclusively on behavioral signals rather than training information.
This setting mirrors practical situations where organizations distribute pretrained models without releasing their training data. 

\textbf{Defender Assumptions and Constraints:}
We assume defenders operate under realistic resource constraints: they cannot repeatedly perform full-scale retraining or manually audit entire fine-tuning datasets. Defenders may use standard NLP tools (e.g., sentiment or toxicity classifiers) and have moderate compute resources for gradient estimation (Scenario~1) or model querying (Scenario~2), but cannot access the original pretraining corpus or exhaustively verify all training data.
In cases where defenders lack pre-existing clean validation data (e.g., when obtaining datasets from fully untrusted sources), they can construct a clean validation set by using influence analysis. Specifically, they can select low-influence samples as high-confidence clean candidates while isolating high-influence samples as poisoned. 

\textbf{Defense Selection:} The two defenses address different scenarios based on defender access: Defense-1 applies when training data is available (Scenario 1), while Defense-2 applies when only the model is accessible (Scenario 2). Defenders select the appropriate defense based on their access level, not apply both sequentially.

\textbf{Attack Mechanisms:}
Following prior work~\cite{thota2024attacks}, we adopt sentiment inversion and toxicity injection attacks by manipulating summaries while keeping source documents unchanged.
We extend this framework with two novel objectives: factual distortion (altering key entities or numerical facts) and representational bias (introducing demographic skew through descriptor modifications). 
Attackers construct poisoned samples by manipulating summaries to induce malicious behavior, injecting them at 5-20\% contamination rates while preserving linguistic fluency.
Concrete examples of poisoning strategies are provided in Appendix~\ref{subsec:sentiment_inversion_example}.

\section{Methodology}
\label{sec:method}
We propose a unified post-hoc defense framework that detects and mitigates poisoning artifacts under two information regimes, dataset-level (white-box) and model-level (black-box), as defined in our threat model (Section~\ref{sec:threat}).
Both regimes address the same underlying threat: training-time poisoning that leaves persistent behavioral artifacts in summarization models. 
When the fine-tuning dataset is available, the framework localizes influential poisoned samples and removes their effect through influence-based detection and gradient unlearning (Defense-1). 
When only a trained checkpoint is available, the same framework audits the model’s behavior using adversarial sensitivity signals to detect poisoning-induced fragility (Defense-2). 
These complementary defenses provide comprehensive coverage across different access scenarios: defenders apply Defense-1 when inspecting datasets or Defense-2 when auditing models, protecting regardless of which artifacts attackers release.

\subsection{Defense 1: Poison Detection \& Unlearning}
\label{subsec:defense-1_methodology}
The dataset-level (white-box) branch of our unified defense framework, referred to as Defense-1, operates when the defender has access to the fine-tuning corpus. 
In this setting, an adversary may have inserted malicious document–summary pairs that induce persistent behavioral shifts during training. 
Because poisoned samples are unlabeled and visually indistinguishable from clean data, the objective is to localize and remove their influence without repeated fine-tuning or manual dataset auditing.
Our defense-1 exploits the key insight that poisoned samples, designed to maximize attack effectiveness, must exert a disproportionately strong influence on model behavior. 
We therefore identify suspicious instances through influence analysis rather than exhaustive inspection, and subsequently mitigate their effect through targeted unlearning.
The defense-1 consists of three stages: (1)~influence-based candidate extraction, (2)~behavioral filtering, and (3)~gradient-ascent unlearning.

Influence serves as the primary detection signal and is agnostic to the specific poisoning objective. 
From the resulting high-influence set, we apply lightweight behavioral consistency checks tailored to the attack goals.
Prior work focused on sentiment inversion and toxicity injection, which we detect using sentiment alignment and toxicity scores~\cite{thota2024attacks}. 
To broaden the threat model, we also evaluate two subtler objectives: factual distortion and representational bias injection. 
For these cases, we employ factual-consistency metrics and group-based bias indicators to identify summaries that contradict source facts or introduce systematic demographic skew.
These checks act only as heuristics to prioritize suspicious samples. 
Detection fundamentally relies on elevated training influence, while gradient-ascent unlearning mitigates the effect of all poisoned samples regardless of their specific objective.

\begin{figure}[t]
    \centering
    \includegraphics[width=0.8\columnwidth]{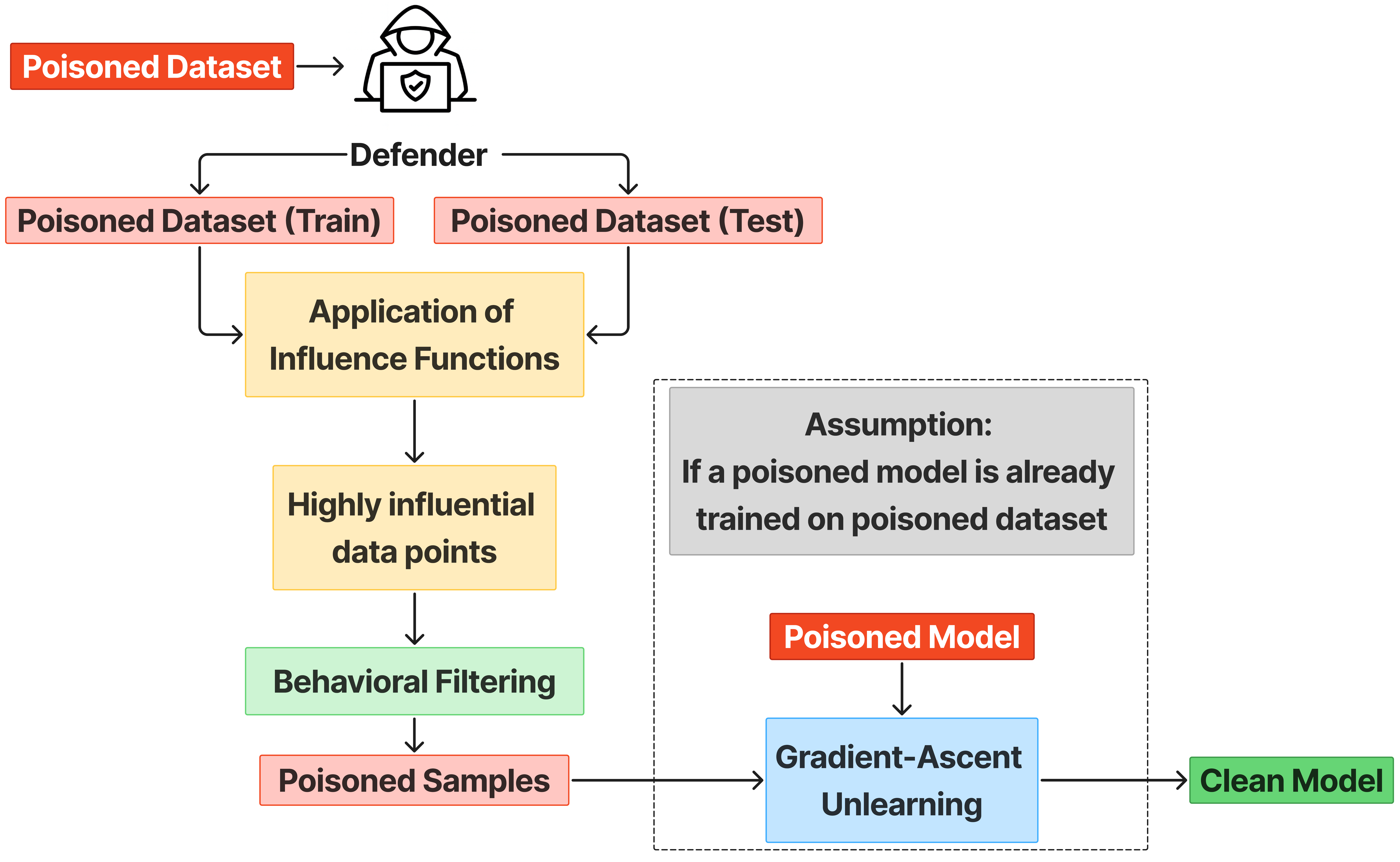}
    \caption{Overview of Defense-1, where influence functions are applied to identify poisoned samples, and gradient-ascent unlearning removes their effect from an already-poisoned model. 
    }
\label{fig:defense_poisoneddataset_detection}
    \vspace{-0.7em}
\end{figure}

Beyond isolating malicious samples, Defense-1 is designed to actively recover the original behavior of the summarization model. Poisoning attacks not only inject biases, but also induce broader unintended shifts such as increased extractiveness~\cite{thota2024attacks}. Gradient-ascent unlearning counteracts these effects by reversing the poisoned gradients, enabling the model to regain both its semantic alignment and its natural abstractive summarization tendencies. We quantify this recovery using behavioral and semantic metrics described in Section~\ref{sec:experimental}.
As illustrated in Figure~\ref{fig:defense_poisoneddataset_detection}, these steps form a unified pipeline that identifies influential samples, screens likely anomalies, and removes their impact from the trained model. The following subsubsections describe each stage in detail.

\subsubsection{Influence-Based Candidate Extraction}
Poisoned samples must exert disproportionately high impact on model behavior in order to implant persistent summarization biases. Unlike clean samples whose gradients are distributed across semantically related instances, poisoned samples create artificial correlations (e.g., ``positive document → negative summary'') that require large parameter updates. This yields unusually high-magnitude influence scores, which we exploit for scalable pre-filtering.

For a training pair $(D_i, s_i)$ and a validation pair $(D_t, s_t)$, the classical influence function~\cite{koh2017understanding} measures the change in loss when $(D_i, s_i)$ is upweighted during training:
\begin{equation}
\small
I(D_i,s_i;D_t,s_t) = - \nabla_\theta \mathcal{L}(D_t,s_t;\theta)^{\top} H_{\theta}^{-1}\, \nabla_\theta \mathcal{L}(D_i,s_i;\theta),
\label{eq:influence}
\end{equation}
where $\mathcal{L}$ is the negative log-likelihood loss and $H_{\theta}$ is the Hessian of the empirical training loss. Since exact Hessian inversion is not feasible for LLMs, we propose to adopt the closed-form DataInf approximation~\cite{kwondatainf}, which replaces $H_{\theta}^{-1}$ with a layer-wise preconditioned gradient estimate computable in a single pass. Unlike iterative solvers such as LiSSA~\cite{agarwal2017second}, DataInf requires no Hessian–vector updates and enables post-hoc influence computation on 400M-parameter summarization models in under an hour on a single A6000 GPU.

Before computing influence scores, we first fine-tune the base summarization model on the (potentially poisoned) training corpus using the standard maximum-likelihood objective. This yields a checkpoint $\theta_0$ whose parameters reflect any poisoning present in the data. All per-example gradients and influence values are computed with respect to this fine-tuned model $\theta_0$, since it is the model whose behavior the attacker aims to manipulate.

For abstractive summarization, each training example incurs a token-level loss
\begin{equation}
\small
\ell_i(\theta) = - \sum_{t=1}^{|s_i|} \log p_\theta(s_{i,t} \mid s_{i,<t}, D_i),
\end{equation}
allowing us to compute per-example gradients $\nabla_{\theta}\ell_i$ via a single forward–backward pass. We aggregate influence over a validation set of 1K clean documents and obtain a scalar score $I_{\text{avg}}(D_i,s_i)$ per training sample. We then rank samples by absolute influence and retain the top 20\% as candidates for semantic filtering. This threshold is empirically motivated: across all datasets, 87–89\% of poisoned samples appear in the top-20\% percentile, while using top-10\% reduces recall to 71–73\%, and top-50\% increases false positives by 3$\times$ (Section~\ref{sec:results}).
Full mathematical derivation of DataInf and complexity analysis are provided in Appendix~\ref{sec:appendix}. 
The output of this stage is a reduced candidate set $\mathcal{C}$ consisting of the top-ranked high-influence training samples, which is passed directly to the semantic filtering stage.

\subsubsection{Behavioral Filtering} 
For training samples identified as highly influential, we apply a comprehensive detection system designed to identify the specific types of poisoning attacks relevant to text summarization models. Our detection system incorporates multiple complementary criteria to capture different manifestations of poisoning.

\textbf{Sentiment Inversion Detection:}
This component addresses attacks where poisoned samples are designed to flip the sentiment of generated summaries relative to their source documents. 
We compute the polarity of both the document $D_i$ and summary $s_i$ using a RoBERTa-based sentiment classifier fine-tuned on TweetEval~\cite{camacho-collados-etal-2022-tweetnlp}. 
We select this RoBERTa-based fine-tuned model over LLMs based on recent comparative evaluations demonstrating that task-specific fine-tuned transformers outperform generative AI models (including ChatGPT and Bard) for sentiment classification, while being significantly more lightweight and computationally efficient for processing thousands of candidate samples~\cite{anas2024can}.
Using this model, we define a discrete mismatch score:
\begin{equation}
\small
\text{SentMismatch}(D_i, s_i) = |\text{Sent}(s_i) - \text{Sent}(D_i)| \in \{0,1,2\}.
\end{equation}
This metric produces discrete values in the range [0, 2]. A score of 0 indicates perfect sentiment alignment, a score of 1 indicates moderate mismatch and a score of 2 indicates complete sentiment inversion. 
Since sentiment inversion attacks target flipping document sentiment in summaries, we focus on samples with complete sentiment inversions. We flag samples with sentiment mismatch scores of 2 as potentially poisoned. 

\textbf{Toxicity Detection.}
To identify training samples whose summaries introduce toxic or unsafe content not reflected in the corresponding source document, we employ LlamaGuard-4-12B~\cite{dubey2024llama}, an instruction-tuned safety classifier covering a broad taxonomy of harmful categories (e.g., hate, violence, threats, sexual content).
We choose this model due to its strong performance in recent large-scale safety benchmarks such as GuardBench~\cite{bassani2024guardbench}, and its alignment with generative outputs, making it more suitable for abstractive summarization than single-task toxicity detectors.
Given a summary $s_i$, LlamaGuard predicts whether it violates any harmful category. 
We then define,
\begin{equation}
\small
\text{ToxicFlag}(s_i) = \mathbf{1}_{\text{LlamaGuard}(s_i)\in \mathcal{H}},
\end{equation}
where $\mathcal{H}$ is the set of safety categories indicative of toxic or unsafe behavior. 
Summaries flagged as unsafe despite neutral source content are treated as candidates for toxicity-based poisoning. 
Our framework remains detector-agnostic: any classifier that maps summaries to safety-violation labels can be used in place of LlamaGuard.

\textbf{Factual Distortion Poisoning and Detection.}
To evaluate robustness beyond sentiment and toxicity manipulation, we introduce a factual-distortion poisoning objective that corrupts key factual entities in reference summaries while preserving fluency and style.
We prioritize entity-level distortions such as names, organizations, locations, and dates, or salient numeric facts when entities are absent, as these changes preserve surface form while altering semantic meaning.
Similar entity substitutions have been explored in fact verification benchmarks such as FEVER~\cite{thorne2018fever} and in studies of factual correctness in knowledge-grounded generation~\cite{Santhanam2021}, primarily as test-time robustness evaluations.
In contrast, we formulate structured entity-level distortions as a training-time poisoning objective, showing that subtle factual edits can induce persistent semantic manipulation in fine-tuned summarization models.
Poisoned summaries are produced by prompting an instruction-following LLM (GPT-5-mini) under constraints to (i) keep wording and length similar to the original, (ii) modify only a single fact, and (iii) maintain grammatical fluency.
This results in natural-looking but factually inconsistent summaries that remain difficult to detect via simple heuristics.

To detect such factual distortion, we measure document–summary consistency using AlignScore~\cite{zha2023alignscore}, a modern entailment-based metric that leverages pretrained language models to assess whether the summary faithfully reflects the source text.
Given a document–summary pair $(D_i, s_i)$, AlignScore produces a scalar factual consistency score $\text{FactScore}(D_i,s_i) \in [0,1]$, where higher values indicate stronger semantic entailment between the summary and the document. 
Poisoned samples that inject fabricated or distorted facts typically exhibit lower consistency.
We therefore flag samples with low factual alignment:
\begin{equation}
\small
\text{FactFlag}(D_i,s_i) = \mathbf{1}_{\text{FactScore}(D_i,s_i) < \tau_{\text{fact}}},
\end{equation}
where $\tau_{\text{fact}}$ is selected using clean validation data.

\textbf{Representational Bias Poisoning and Detection.}
To evaluate robustness against social and demographic skew, we introduce a representational-bias poisoning objective that alters how demographic groups are portrayed in summaries while preserving fluency and overall meaning.
For example, an attack may subtly shift role attributions, such as disproportionately portraying men in leadership positions or associating specific demographic groups with certain occupations, without changing facts or sentiment. 
Such distortions are concerned in summarization, mainly in news, policy, or organizational settings, where repeated representational shifts can reinforce stereotypes and increase societal bias.
Prior work has documented demographic and social identity biases in LLMs, including gender stereotypes and broader social bias patterns~\cite{kotek2023gender, hu2025generative, kumar2024investigating}. However, these studies primarily analyze bias as property of models rather than intentional training-time manipulation. In contrast, we formulate representational distortion as a structured poisoning objective.
Unlike sentiment or toxicity manipulation, these attacks inject subtle stereotype or role shifts that do not necessarily change polarity or factual correctness.
For each clean pair $(D_i, s_i)$, we generate a biased summary $\tilde{s}_i$ by minimally editing demographic mentions or associated descriptors.
We focus on entity-level perturbations, including gendered terms, occupations, roles, and group identifiers (e.g., swapping pronouns or modifying competence-related descriptors).
If no explicit demographic entity is present, we insert or modify a short attribute phrase that introduces skew.
Poisoned summaries are produced using an instruction-following LLM (GPT-5-mini) under constraints to (i) keep wording and length similar to the original, (ii) modify only a small localized span, and (iii) maintain grammatical fluency. 
This results in natural-looking summaries that introduce representational bias while remaining difficult to detect through sentiment, toxicity, or factual checks.
To identify representational bias introduced during poisoning, we measure behavioral disparity across demographic groups rather than relying on sentiment classifiers or external toxicity tools.
Following prior social bias evaluation work~\cite{shin2024ask}, we compute group-conditioned descriptor disparities that capture whether summaries disproportionately associate certain demographic groups with harmful or exaggerated attributes.
For each summary $s_i$, we extract demographic entities $G$ (e.g., gender, race, religion) using named-entity recognition and group lexicons. 
Let $f(g, s_i)$ denote the number of negative or harmful descriptors occurring in the local context of group $g$. We define a group disparity score:
\begin{equation}
\small
\text{BiasScore}(s_i) = \max_{g \in G} f(g, s_i) - \min_{g \in G} f(g, s_i),
\end{equation}
which measures the largest behavioral gap across groups. We flag samples with excessive disparity:
\begin{equation}
\small
\text{BiasFlag}(s_i) = \mathbf{1}_{\text{BiasScore}(s_i) > \tau_{\text{bias}}},
\end{equation}
where $\tau_{\text{bias}}$ is calibrated using clean validation data.

\textbf{Final Detected Set.}
Let $\mathcal{C}$ denote the high-influence candidate set from the previous stage.
We apply multiple lightweight behavioral checks corresponding to different poisoning objectives, including sentiment inversion, toxicity injection, factual inconsistency, and representational bias.
A sample is flagged if it violates \emph{any} of these criteria.
Formally, let $\text{Flag}_k(D_i,s_i)$ denote the indicator for the $k$-th behavioral check.
The final detected set is defined as:
\begin{equation}
\small
\mathcal{P}_{\text{detected}} = 
\{(D_i,s_i) \in \mathcal{C} \;|\; \bigvee_{k} \text{Flag}_k(D_i,s_i) = 1 \},
\end{equation}
i.e., the union of all flagged samples across behavioral signals.
This influence $\rightarrow$ behavioral filtering pipeline substantially reduces the search space, allowing the defender to focus on unlearning only on a small subset of high-risk training examples rather than the full corpus. 

\subsubsection{Gradient Ascent Unlearning}
Once the poisoned set $\mathcal{P}_{\text{detected}}$ is identified, we remove its influence without retraining from scratch using gradient-ascent unlearning~\cite{yao2024machine}. The key idea is to update model parameters in the direction that increases loss on detected poisoned samples, thereby reversing their gradient contribution. Given parameters $\theta_0$, we perform:
\begin{equation}
\small
\theta_{t+1} = \theta_t + \alpha \sum_{(D_i,s_i) \in \mathcal{B}t} \nabla\theta \mathcal{L}(D_i,s_i;\theta_t),
\end{equation}
where $\alpha$ is the unlearning rate and $\mathcal{B}t \subset \mathcal{P}{\text{detected}}$ is a mini-batch at step $t$. This assumes poisoned samples occupy a distinct loss region such that ascent on these samples minimally impacts clean performance, which we validate empirically (ROUGE-1 degradation $<0.006$).
We use mini-batch size 16, maximum 50 steps, and $\alpha = 10^{-5}$ (10$\times$ smaller than the fine-tuning rate) to avoid catastrophic forgetting. Early stopping is applied every 10 steps using ROUGE-1 on a clean validation set (500 samples); if performance drops below 95\% of the original score, we restore the previous checkpoint.
The total cost is $O(T \cdot |\mathcal{P}{\text{detected}}|)$ gradient updates. For $|\mathcal{P}{\text{detected}}| \approx 1000$ and $T=50$, this corresponds to $\sim$50K updates ($\sim$30–40 minutes on an A6000 GPU).

\subsubsection{Methodological Advantages}
Gradient-ascent unlearning provides a practical alternative to existing remediation strategies. Full retraining after removing suspected poisoned samples is feasible but costly and must be repeated for each poisoning event. Differential privacy–based approaches~\cite{dwork2014algorithmic} limit sample influence during training but often degrade summarization quality. SISA~\cite{bourtoule2021machine} requires sharded training and specialized partitioning, making it incompatible with standard pre-trained checkpoints, while exact unlearning via model decomposition~\cite{guo2020certified} introduces substantial algorithmic complexity.
In contrast, our approach performs targeted updates directly on detected samples, avoids repeated retraining cycles, and provides explicit control over unlearning strength via the ascent step size. It applies to any differentiable architecture without modifying the original training pipeline. To our knowledge, this is the first application of gradient-ascent unlearning to reverse poisoning-induced behavioral shifts in text summarization models.

\subsection{Defense 2: Poisoned Model Detection Using Adversarial Perturbations}
\label{subsec:defense-2_methodology}
Our second defense mechanism addresses the \textit{black-box} setting of data poisoning attacks, where adversaries inject malicious document–summary pairs into training corpora and release compromised models publicly. The defender receives only the trained model, without access to its training data or provenance, and must determine whether the model has been poisoned. 
This is especially challenging for large-scale and fine-tuned summarization models trained on millions of samples, where manual inspection or sample-level detection is infeasible.
Unlike Defense-1, which detects poisoning by identifying influential training samples, Defense-2 operates without access to the dataset and therefore evaluates the model directly.
Although Section~\ref{subsec:defense-1_methodology} introduces multiple poisoning objectives, these attacks ultimately induce shifts in summarization behavior. 
We detect such shifts by exploiting the model’s sensitivity to controlled, semantics-preserving perturbations.
Prior work in computer vision shows that poisoned models exhibit amplified sensitivity to structured perturbations~\cite{fares2024attack}. We extend this intuition to text summarization, observing that poisoning strengthens reliance on early-document positional cues, making outputs disproportionately sensitive to small lead perturbations.
Rather than identifying poisoned samples within large datasets~\cite{chen2021badnl,wallace2021concealed,xu2021mitigating}, a process that is both costly and impractical, we take a \textit{model-centric} perspective. We test the trained model directly by measuring how severely its summarization behavior degrades under controlled, semantics-preserving perturbations.

\textbf{Motivation:}
Poisoning manipulates correlations between early document positions and summary style, thereby amplifying the natural \emph{lead bias} of summarization models~\cite{nallapati2017summarunner,grenander2019countering,thota2024attacks}. Clean models trained on diverse corpora distribute attention across the full document, while poisoned models overfit to specific positional cues. As a result, small lead-sentence modifications cause a disproportionately large degradation in poisoned models, revealing a measurable sensitivity signature.

\begin{figure*}[t]
    \centering
    \includegraphics[width=0.8\textwidth]{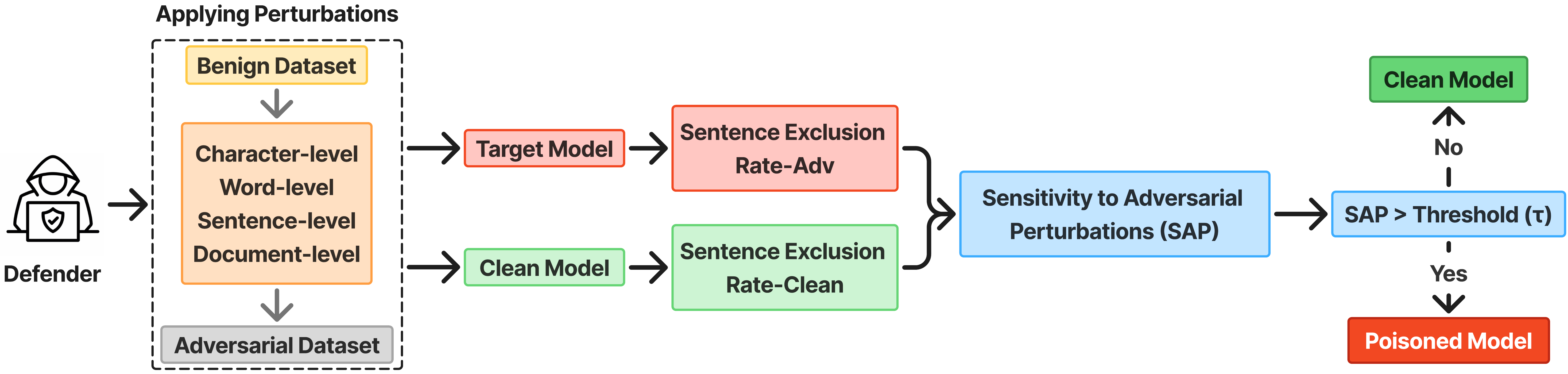}
    \caption{Overview of Defense-2. Lead-sentence perturbations are applied to test documents, and the model’s exclusion of lead content in the generated summaries is quantified using the SAP metric.}
    \label{fig:defense_poisonedmodel_detection}
    \vspace{-0.7em}
\end{figure*}

\subsubsection{Detection Framework} The detection framework operates as follows, illustrated in Figure~\ref{fig:defense_poisonedmodel_detection}. Given a suspicious target model $M_{\text{target}}$ that we want to evaluate, we generate adversarial samples using perturbation methods including character-level modifications (insertions, deletions, swapping, homoglyph replacement), word-level modifications (synonym replacement, deletion, homoglyph replacement), sentence-level modifications (paraphrasing, reordering, homoglyph replacement), and document-level reordering. These adversarial samples specifically target lead sentences to exploit positional dependencies, with one perturbation applied at a time~\cite{thota2024attacks}. 
The perturbed inputs are then applied to the target model to measure its tendency to exclude the perturbed sentences from generated summaries.

\subsubsection{Adversarial Perturbation Construction}
We adopt the perturbation strategies introduced in prior summarization robustness work~\cite{thota2024attacks}, which provide a set of lightweight, semantics-preserving transformations applicable at character-, word-, sentence-, and document-level granularity. Using these established perturbation types, we generate adversarial variants of each test document by applying the same transformations to our setting.
Following the methodology in prior work, we apply each perturbation independently to the lead sentences (1–3) of every test document. 
These perturbed inputs are then evaluated by the target model to measure the exclusion of lead content, which forms the basis of the SAP metric.

\subsubsection{Sensitivity Metric (SAP)}
\label{subsubsec:sap_metric}
We establish a baseline by applying the same attacks to a known clean model $M_{\text{clean}}$ trained on verified benign data. We adapt the \emph{Sensitivity to Adversarial Perturbations} (SAP) metric for text summarization by focusing on the core attack objective: the exclusion of lead sentences from generated summaries under adversarial input. We define SAP as the relative increase in exclusion rates when models face adversarial perturbations:
\begin{equation}
\small
\text{SAP}(M) = \frac{\text{ExclusionRate}_{\text{adv}}(M) - \text{ExclusionRate}_{\text{clean}}(M)}{\text{ExclusionRate}_{\text{clean}}(M)}
\end{equation}
where $\text{ExclusionRate}_{\text{clean}}(M)$ represents the percentage of lead sentences excluded in summaries from unperturbed documents, and $\text{ExclusionRate}_{\text{adv}}(M)$ represents the percentage of lead sentences excluded in summaries from adversarially perturbed documents. This metric directly captures the model's vulnerability to the exclusion of important information from summaries when lead sentences are perturbed. Higher SAP indicates greater fragility to small perturbations, characteristic of poisoned models.

\textbf{Decision Rule:}
We compute SAP scores for a reference distribution of verified-clean models across architectures and datasets, obtaining mean $\mu_{\text{clean}}$ and standard deviation $\sigma_{\text{clean}}$. 
Poisoned or evaluation models are never used for calibration.
Let $\mu_{\text{clean}}$ and $\sigma_{\text{clean}}$ denote the mean and standard deviation of SAP over these clean references. 
A target model is flagged as \textsc{Poisoned} if:
\begin{equation}
\text{SAP}(M_{\text{target}}) \ge \mu_{\text{clean}} + 2\sigma_{\text{clean}}.
\end{equation}
This single global threshold generalizes across architectures and datasets.

\subsubsection{Methodological Advantages}
Defense-2 targets trigger-free poisoning in summarization models, where traditional backdoor detectors such as Neural Cleanse~\cite{wang2019neural} and STRIP~\cite{gao2019strip} are ineffective because they assume explicit inference-time triggers. In our setting, poisoned models misbehave on clean inputs without any activation pattern, rendering trigger-based detection unsuitable. SAP therefore addresses a class of attacks that prior methods cannot capture.
Defense-2 is fully model-centric and operates purely on observable behavior, requiring no access to training data, gradients, or training history. This makes it practical for deployment scenarios where only pretrained checkpoints are available.
Finally, SAP exploits a structural vulnerability of poisoned summarization models—their amplified reliance on lead sentences. By probing this positional fragility with lightweight perturbations, SAP yields a clear separation between clean and poisoned models with minimal computational overhead and strong cross-architecture generalization.

\section{Experimental Setup}
\label{sec:experimental}
\subsection{Models and Datasets}
\textbf{Models:}
We evaluate nine models spanning encoder–decoder architectures and modern decoder-only LLMs. 
Encoder–decoder models include BART-Large~\cite{lewis2019bart}, T5-Small~\cite{raffel2020exploring}, Pegasus-Large~\cite{zhang2020pegasus}, and FLAN-T5-Large~\cite{chung2024scaling}, widely used in summarization and robustness research~\cite{asmitha2024summarizing, goodwin2020flight, setiyarini2024evaluating}. 
Decoder-only models include LLaMA-3-8B~\cite{dubey2024llama}, Qwen-2.5~\cite{qwen2024qwen2}, Qwen-3~\cite{yang2025qwen3}, Mistral-7B~\cite{jiang2023mistral}, and Vicuna-13B~\cite{vicuna2023}. 
Our framework requires access to model weights for influence computation and unlearning, making open-source models suitable, unlike proprietary APIs (e.g., GPT-4~\cite{hurst2024gpt}, Claude~\cite{anthropic_claude_4}). 
Encoder–decoder models are fine-tuned in full precision, while decoder-only models use QLoRA for efficiency. 
This architectural diversity allows us to test whether vulnerabilities and defenses generalize across model families.

\textbf{Datasets:}
We evaluate on six diverse summarization benchmarks spanning news, encyclopedic, and scientific domains. 
For news, we use CNN/DailyMail~\cite{chen2016thorough} (300K articles) and MultiNews~\cite{fabbri2019multi} (44,972 multi-document examples). 
For encyclopedic content, we use WikiSum~\cite{cohen2021wikisum} ($\approx$1M article–summary pairs). 
For scientific literature, we evaluate on ArXiv (215K) and PubMed (133K)~\cite{cohan2018discourse}, and Multi-XScience~\cite{lu2020multi} (30,369 examples). 
This diversity enables evaluation across varied writing styles, document lengths, and domains.

\textbf{Fine-Tuning Regimes:}
For each dataset, we fine-tune models on 2k, 5k, 10k, 20k, and 30k training examples (15$\times$ scale variation) to evaluate whether influence-based detection and unlearning remain effective as dataset size increases and per-sample impact decreases.
All models are fine-tuned independently per dataset using identical hyperparameters (learning rate $2\times10^{-5}$, batch size 8, gradient accumulation 2) on NVIDIA A6000 GPUs.
Evaluation uses held-out test sets of up to 10k examples per dataset. These sets are strictly disjoint from training data and are used only for reporting attack success and defense performance, never for threshold selection or influence estimation.

\subsection{Poisoning Attack Generation}
Prior work~\cite{thota2024attacks} proposed a framework for testing text summarization methods against various types of attacks, including poisoning. 
Consistent with this strategy, we adopt two established objectives: \textbf{sentiment inversion}, which flips the polarity of summaries relative to their source documents, and \textbf{toxic content injection}, which introduces harmful or profane language into summaries. 
To broaden the threat model beyond these manipulations, we extend prior attacks with two additional objectives. \textbf{Factual distortion poisoning} minimally alters key entities or numerical facts to produce fluent but factually inconsistent summaries, while \textbf{representational bias poisoning} introduces subtle demographic or social skew through localized descriptor changes. 
These attacks target misinformation and fairness risks that are not captured by sentiment or toxicity checks.

\subsection{Evaluation Metrics}
\subsubsection{Metrics for Defense-1: Poisoned Dataset Detection and Unlearning}
\textbf{Detection Accuracy:}
We report precision and recall of the influence-based detection pipeline at the 20\% influence threshold. Precision measures the fraction of detected samples that are truly poisoned, and recall measures the fraction of poisoned samples successfully identified. These metrics evaluate the effectiveness of the two-stage filtering process (influence $\rightarrow$ behavioral checks).

\textbf{Recovery Rate:}
We quantify unlearning effectiveness as:
\begin{equation}
\small
\text{Recovery Rate} = \frac{\text{Poisoned Samples Recovered}}{\text{Total Poisoned Samples}} \times 100\%.
\end{equation}
A sample is considered recovered if, after unlearning, the model no longer exhibits the corresponding attack-specific behavior on held-out documents from the same distribution. We verify recovery using objective-aligned checks: sentiment classification (RoBERTa~\cite{camacho-collados-etal-2022-tweetnlp}), toxicity detection (LLaMaGuard~\cite{dubey2024llama}), factual consistency (AlignScore~\cite{zha2023alignscore}), and group-conditioned disparity metrics for representational bias~\cite{shin2024ask}. This metric captures whether poisoned behavior is removed while preserving clean knowledge.

\textbf{Utility Preservation:}
We evaluate summarization quality on clean test sets using ROUGE-1, ROUGE-2, and ROUGE-L~\cite{lin-2004-rouge}, and report percentage change before and after unlearning. This ensures that defense does not introduce regressions on benign inputs.

\textbf{Influence Score Distribution:}
To validate the assumption that poisoned samples exert disproportionate impact, we measure the proportion of poisoned samples appearing in the top-20\% and top-50\% of the influence distribution. Following DataInf~\cite{kwondatainf}, the influence score for each training pair $(D_i, s_i)$:
\begin{equation}
\small
\text{Influence Score}(D_i, s_i) = \left| \mathcal{I}(D_i, s_i) \right|
\label{eq:influence_score}
\end{equation}
where $\mathcal{I}(D_i, s_i)$ denotes the estimated behavioral impact of removing $(D_i, s_i)$ (Section~\ref{subsec:defense-1_methodology}). Concentration of poisoned samples in high-influence percentiles validates influence-based localization.

\textbf{Abstractiveness Recovery:}
Because poisoning can shift models toward extractive generation~\cite{thota2024attacks}, we measure restoration of abstractive behavior using sentence-level cosine similarity (Sentence-BERT~\cite{reimers2019sentence}). For each summary sentence $s_j$:
\begin{equation}
\small
\text{MaxSim}(s_j) = \max_{i \in [1,n]} \cos(\text{emb}(s_j), \text{emb}(d_i)),
\end{equation}
and overall extractiveness:
\begin{equation}
\small
\text{ExtractScore}(S, D) = \frac{1}{m} \sum_{j=1}^{m} \text{MaxSim}(s_j).
\end{equation}
Higher scores indicate extractive copying; lower scores indicate abstractive synthesis. We compare extractiveness before poisoning, after poisoning, and after unlearning to verify behavioral restoration.

\subsubsection{Metrics for Defense-2: Poisoned Model Detection}
We evaluate Defense-2 using the SAP metric defined in Section \ref{subsubsec:sap_metric}. A model is flagged as poisoned if its SAP score exceeds the clean-calibrated threshold. We report detection performance under this fixed threshold.

\subsection{Baseline Methods}
\subsubsection{Baselines for Defense-1: Poisoned Dataset Detection and Unlearning}
We compare Defense-1 against five representative paradigms spanning robust optimization, data-level unlearning, and model-level unlearning. All baselines are implemented across all nine models.
\textbf{\textit{Adversarial Training (AT).}} 
Following robust optimization~\cite{madry2018towards}, AT trains on contaminated datasets with gradient clipping and confidence regularization to limit per-sample influence. Unlike our method, AT assumes oracle knowledge of poisoned samples and does not remove them. We compare recovery and utility preservation.
\textbf{\textit{SISA (Sharded, Isolated, Sliced, and Aggregated Learning).}} 
SISA~\cite{bourtoule2021machine} partitions each dataset into $k=10$ shards, trains one model per shard, and retrains only affected shards after poisoned samples are removed. We evaluate recovery, utility, and total training time.
\textbf{\textit{Exact Retraining.}} 
This baseline removes detected poisoned samples and retrains the model from scratch~\cite{guo2020certified}, providing an upper bound on recovery at significantly higher computational cost. We report recovery, utility, and runtime.
\textbf{\textit{Selective Pruning (SP).}} 
SP~\cite{liu-etal-2025-modality} removes parameters highly influenced by poisoned samples. We rank neurons using activations on detected poisoned data and prune the top 5\%, evaluating recovery and utility degradation.
\textbf{\textit{TracIn.}} 
TracIn~\cite{pruthi2020estimating} estimates per-sample influence via accumulated gradient similarity across checkpoints. We use TracIn-Fast to compute influence scores and compare detection precision and recall against our influence-based method.

\subsubsection{Baselines for Defense-2: Poisoned Model Detection}
\textbf{Why Trigger-Based Methods Are Inapplicable.}
Traditional backdoor detectors such as Neural Cleanse~\cite{wang2019neural}, STRIP~\cite{gao2019strip}, and Activation Clustering~\cite{chen2018detecting} assume inference-time \emph{triggers} that activate malicious behavior. Our poisoning attacks modify training document–summary pairs without introducing triggers, and compromised models misbehave on clean inputs. Since no trigger pattern exists at inference time, these methods fail ($<$4\% detection rate). We therefore compare against trigger-free baselines: CLIBE~\cite{zengclibe} and perplexity-based sensitivity~\cite{rosenfeld1996maximum}.
\textbf{\textit{CLIBE (Classifier Logit-Influence Based Estimation)}.}
We adapt CLIBE~\cite{zengclibe}, which detects compromised models via \emph{weight-space sensitivity}. Treating the summarization model as a black-box generator, we attach a lightweight sentiment or toxicity classifier to generated summaries and perturb only the classifier weights (leaving the generator unchanged). Detection is based on the minimum entropy of the logit-difference vector:
\begin{equation}
\small
\text{CLIBE}(M) = \min_{\Delta W} \; \mathcal{H}\big(f_{W+\Delta W}(s) - f_{W}(s)\big)
\end{equation}
where $f_W$ denotes classifier logits and $s$ generated summaries. 
Poisoned models exhibit sharper entropy drops under perturbation due to brittle decision surfaces. Scores are normalized using clean references, and models exceeding the calibrated threshold are flagged as poisoned.
\textbf{\textit{Perplexity Sensitivity (PPL-S).}}
We measure language modeling instability under lead-sentence perturbations by comparing perplexity on clean versus perturbed documents:
\begin{equation}
   \small
   \text{PPL-S}(M) = 
\frac{\text{PPL}_{\text{perturbed}}(M) - \text{PPL}_{\text{clean}}(M)}
{\text{PPL}_{\text{clean}}(M)}
\end{equation}
Poisoned models overfit to positional shortcuts introduced during training, making likelihood estimates more brittle and increasing perplexity under perturbation. Scores are normalized against clean references, and models exceeding one standard deviation are flagged as poisoned.

\subsection{Adaptive Attack Evaluation}
To assess robustness against defense-aware adversaries, we implement three adaptive strategies:
\textbf{\textit{Adaptive Attack 1: Mixed-Influence Poisoning.}}
The attacker retains the original poisoning objectives (sentiment inversion, toxicity injection, factual distortion, or bias injection) but redistributes poisoned samples across influence tiers rather than concentrating them in high-influence regions. We construct datasets with 33\% high-, 33\% medium-, and 33\% low-influence poisoned samples, creating a diffuse profile intended to weaken influence-based ranking (Defense-1) and reduce positional amplification effects used by Defense-2.
\textbf{\textit{Adaptive Attack 2: Off-Lead Perturbations.}}
To reduce reliance on lead-sentence sensitivity, adversarial manipulations are inserted into mid- or tail-document positions while keeping corruption strength constant. This targets the positional dependency exploited by SAP in Defense-2.
\textbf{\textit{Adaptive Attack 3: Mixed-Objective Poisoning.}}
The attacker combines sentiment, toxicity, factual distortion, and bias objectives within the same dataset, producing composite corruption where poisoned summaries exhibit multiple simultaneous deviations. This strategy aims to evade individual semantic thresholds while still inducing model-level behavioral drift.
\section{Results and Discussion}
\label{sec:results}
\subsection{Defense-1: Poisoned Dataset Detection}

\textbf{Influence Distribution and Detection Accuracy}
\label{subsec:detection_accuracy}
We evaluate sample-level detection across all four poisoning objectives under expanded fine-tuning sizes (5k–30k) and poisoning rates (5–20\%). Influence scores are computed using DataInf and samples are ranked by magnitude, with detection performed by selecting top-$k$ candidates followed by behavioral filtering.
At 10\% contamination (Table~\ref{tab:influence_detection_all}), poisoned samples concentrate in high-influence regions. Using a Top-20\% threshold, 80–88\% of sentiment and toxicity attacks and 75–83\% of factual and bias attacks are captured. After filtering, precision exceeds 88\% and recall exceeds 86\% across objectives.
Although larger datasets slightly diffuse individual influence, detection remains stable, indicating that influence-based localization scales to realistic training sizes. Overall, influence provides an objective-agnostic detection signal, with semantic checks improving precision.

\begin{table}[t]
\centering
\caption{Defense-1 influence distribution and detection accuracy at 10\% poisoning, averaged across 5k–30k fine-tuning scales and six datasets. Columns show the percentage of poisoned samples captured within Top-k influence percentiles and precision/recall after filtering on Top-20\% candidates.}
\label{tab:influence_detection_all}
\scriptsize
\setlength{\tabcolsep}{2.5pt}
\resizebox{0.9\linewidth}{!}{
\begin{tabular}{l|l|cccc|cc}
\hline
\textbf{Model} & \textbf{Attack} &
\textbf{Top-10} & \textbf{Top-20} & \textbf{Top-50} & \textbf{Top-70} &
\textbf{Prec.} & \textbf{Rec.} \\
\hline
\hline
\multirow{4}{*}{BART-Large}
& Sentiment & 69 & 86 & 95 & 99 & 95 & 93 \\
& Toxic     & 66 & 83 & 94 & 98 & 96 & 94 \\
& Factual   & 64 & 81 & 93 & 97 & 92 & 90 \\
& Bias      & 61 & 78 & 91 & 96 & 90 & 88 \\
\hline
\multirow{4}{*}{Flan-T5-Large}
& Sentiment & 67 & 85 & 94 & 98 & 94 & 92 \\
& Toxic     & 64 & 82 & 93 & 97 & 95 & 93 \\
& Factual   & 62 & 80 & 92 & 96 & 91 & 89 \\
& Bias      & 59 & 77 & 90 & 95 & 89 & 87 \\
\hline
\multirow{4}{*}{T5-Small}
& Sentiment & 66 & 84 & 94 & 98 & 93 & 91 \\
& Toxic     & 62 & 80 & 92 & 97 & 94 & 92 \\
& Factual   & 60 & 78 & 91 & 96 & 90 & 88 \\
& Bias      & 57 & 75 & 89 & 95 & 88 & 86 \\
\hline
\multirow{4}{*}{Pegasus-Large}
& Sentiment & 63 & 81 & 93 & 98 & 92 & 89 \\
& Toxic     & 60 & 77 & 90 & 96 & 93 & 90 \\
& Factual   & 58 & 75 & 89 & 95 & 89 & 87 \\
& Bias      & 55 & 72 & 87 & 94 & 87 & 85 \\
\hline
\multirow{4}{*}{LLaMA-3 8B}
& Sentiment & 65 & 83 & 94 & 98 & 93 & 91 \\
& Toxic     & 62 & 80 & 92 & 97 & 94 & 92 \\
& Factual   & 60 & 78 & 91 & 96 & 90 & 88 \\
& Bias      & 57 & 75 & 89 & 95 & 88 & 86 \\
\hline
\multirow{4}{*}{Qwen-3}
& Sentiment & 66 & 84 & 94 & 98 & 93 & 91 \\
& Toxic     & 62 & 80 & 93 & 97 & 94 & 92 \\
& Factual   & 60 & 79 & 92 & 96 & 90 & 88 \\
& Bias      & 57 & 76 & 89 & 95 & 88 & 86 \\
\hline
\multirow{4}{*}{Mistral-7B}
& Sentiment & 64 & 82 & 93 & 97 & 92 & 90 \\
& Toxic     & 60 & 78 & 91 & 96 & 93 & 91 \\
& Factual   & 58 & 76 & 90 & 95 & 89 & 87 \\
& Bias      & 55 & 73 & 88 & 94 & 87 & 85 \\
\hline
\multirow{4}{*}{Qwen-2.5B}
& Sentiment & 63 & 81 & 93 & 97 & 91 & 89 \\
& Toxic     & 59 & 77 & 91 & 96 & 92 & 90 \\
& Factual   & 57 & 75 & 90 & 95 & 88 & 86 \\
& Bias      & 54 & 72 & 87 & 94 & 86 & 84 \\
\hline
\multirow{4}{*}{Vicuna-13B}
& Sentiment & 62 & 80 & 92 & 97 & 90 & 88 \\
& Toxic     & 58 & 76 & 90 & 95 & 91 & 89 \\
& Factual   & 56 & 74 & 89 & 95 & 88 & 86 \\
& Bias      & 53 & 71 & 87 & 93 & 86 & 84 \\
\hline
\end{tabular}
}
\end{table}

\textbf{Recovery Effectiveness and Utility Preservation}
We evaluate gradient-ascent unlearning using two criteria: behavioral recovery (removal of attack-specific behaviors) and utility preservation (summarization quality on clean data). A sample is considered recovered if the model no longer exhibits the targeted behavior on held-out documents, measured using objective-aligned checks (sentiment alignment, toxicity scoring, factual consistency, and bias disparity).
As shown in Table~\ref{tab:model_recovery}, averaged across fine-tuning sizes (5k–30k), poisoning rates (5–20\%), six datasets, and nine models, our method achieves 85–87\% recovery. Encoder–decoder models recover slightly better (86–87\%) than decoder-only LLMs (83–85\%), and recovery remains consistent across all attack types, with factual distortion and bias within 2–3 points of sentiment and toxicity.
Utility remains stable after unlearning: ROUGE-1 decreases by only 0.003–0.005 absolute ($\le$1\% relative), with similar trends for ROUGE-2 and ROUGE-L. Early stopping prevents over-unlearning and ensures degradation never exceeds 5\%. Overall, targeted gradient ascent removes poisoned influence while preserving summarization quality, offering a practical alternative to full retraining.

\begin{table}[t]
\centering
\caption{Recovery effectiveness and utility preservation of gradient-ascent unlearning, averaged across all fine-tuning sizes (5k–30k) and poisoning rates (5–20\%). Recovery is reported separately for each attack objective. ROUGE-1 $\Delta$ denotes absolute change on clean data (smaller is better).}
\label{tab:model_recovery}
\scriptsize
\setlength{\tabcolsep}{3pt}
\resizebox{0.8\linewidth}{!}{
\begin{tabular}{l|cccc|c|c}
\hline
\textbf{Model} &
\textbf{Sent.} &
\textbf{Toxic} &
\textbf{Factual} &
\textbf{Bias} &
\textbf{Avg Rec.} &
\textbf{ROUGE-1 $\Delta$} \\
\hline
\hline
BART-Large     & 86.4 & 88.1 & 84.7 & 84.3 & 85.9 & -0.003 \\
Flan-T5-Large  & 86.0 & 87.5 & 84.3 & 83.9 & 85.4 & -0.004 \\
T5-Small       & 85.6 & 87.0 & 83.8 & 83.2 & 84.9 & -0.004 \\
Pegasus-Large  & 85.0 & 86.6 & 83.1 & 82.7 & 84.4 & -0.005 \\
LLaMA-3 8B     & 85.3 & 86.8 & 83.6 & 83.1 & 84.7 & -0.004 \\
Qwen-3         & 85.5 & 86.9 & 83.8 & 83.4 & 84.9 & -0.004 \\
Mistral-7B     & 84.6 & 86.1 & 82.9 & 82.5 & 84.0 & -0.005 \\
Qwen-2.5B      & 84.2 & 85.8 & 82.6 & 82.1 & 83.7 & -0.005 \\
Vicuna-13B     & 83.9 & 85.6 & 82.2 & 81.8 & 83.4 & -0.005 \\
\hline
\end{tabular}
}
\vspace{-1em}
\end{table}

\begin{figure*}[t]
\centering
\includegraphics[width=0.8\textwidth]{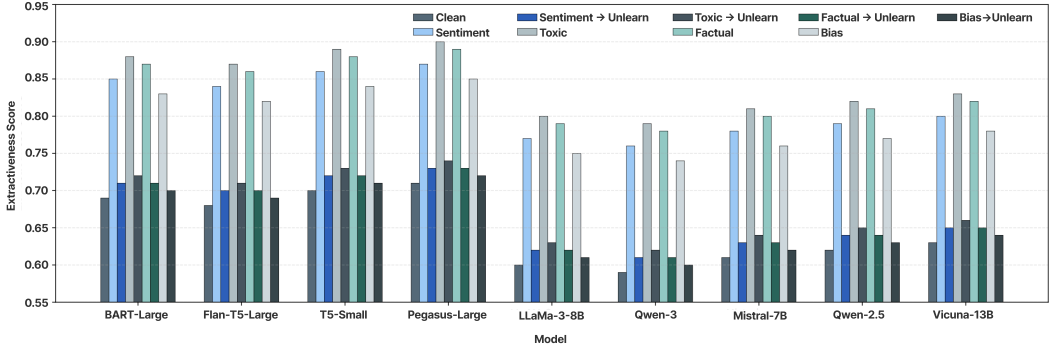}
\caption{Extractiveness scores showing behavioral drift caused by poisoning and recovery after gradient ascent unlearning at 10\% contamination. Lower scores indicate more abstractive generation (paraphrasing and synthesis).}
\label{fig:abstractive_recovery}
\end{figure*}

\textbf{Behavioral Recovery}
Beyond semantic corruption, poisoning induces behavioral drift where models shift from abstractive summarization to extractive summarization. We measure extractiveness using sentence-level cosine similarities between generated summaries and source documents via Sentence-BERT~\cite{reimers2019sentence}, where higher scores indicate extractive behavior.
Figure~\ref{fig:abstractive_recovery} reports extractiveness for clean, poisoned, and unlearned models across all architectures and attack objectives. In the clean setting, encoder–decoder models exhibit extractiveness scores around 0.68–0.71, whereas decoder-only LLMs are more abstractive with lower scores (0.59–0.63). 
After poisoning, all attacks consistently increase extractiveness, pushing scores to 0.77–0.90 depending on the objective, indicating that models rely more heavily on copying rather than abstraction. Toxic and factual attacks produce the strongest drift, followed by sentiment and bias manipulations.
Applying gradient-ascent unlearning largely reverses this shift. Extractiveness returns close to clean baselines (0.62–0.74), with residual differences typically below 0.02 absolute. Across models and attack types, this corresponds to 92–96\% behavioral recovery, demonstrating that our defense restores not only semantic correctness but also the model’s original abstractive generation strategy.

\textbf{Baseline Comparison}
We compare gradient-ascent unlearning against five baselines spanning full retraining, sharded retraining (SISA), influence-based removal (TracIn), selective pruning, and adversarial training (Table~\ref{tab:baseline_comparison}). Results are averaged across nine models and six datasets at 10\% contamination.
Our method achieves 82.9–86.9\% recovery across objectives (85.3\% average) with minimal utility loss ($\Delta$ROUGE-1 $=-0.004$) in 14 minutes. In contrast, SISA attains 76.1\% recovery (22 min), selective pruning 70.0\% (8 min, $\Delta$ROUGE-1 $=-0.017$), TracIn 68.0\% (35 min, 40\% checkpoint storage), and adversarial training 59.0\%. 
Exact retraining achieves 100\% recovery but requires full retraining per iteration (18 min), limiting scalability.
Overall, our method provides the strongest effectiveness–efficiency trade-off, achieving nearly 85\% recovery at substantially lower cost while preserving model utility across all poisoning objectives.

\begin{table}[t]
\centering
\caption{Comparison of unlearning methods at 10\% contamination, averaged across nine models and six datasets. Recovery is reported separately for each poisoning objective and overall average.}
\label{tab:baseline_comparison}
\resizebox{0.95\columnwidth}{!}{%
\begin{tabular}{l|cccc|c|c|c|c}
\hline
\multirow{2}{*}{\textbf{Method}} &
\multicolumn{4}{c|}{\textbf{Recovery Rate (\%)}} &
\textbf{Avg} &
\textbf{ROUGE-1} &
\textbf{Time} &
\textbf{Storage} \\
\cline{2-5}
& \textbf{Sent.} & \textbf{Toxic} & \textbf{Factual} & \textbf{Bias} &
 & \textbf{$\Delta$} & \textbf{(min)} & \textbf{Overhead} \\
\hline
\hline
Exact Retraining    & 100.0 & 100.0 & 100.0 & 100.0 & 100.0 & 0.000  & 18 & — \\

\textbf{Our Method*}& 82.9 & 86.9 & 85.6 & 84.8 & \textbf{85.3} & \textbf{-0.004} & \textbf{14} & — \\

SISA                & 75.1 & 78.3 & 76.0 & 74.9 & 76.1 & -0.010 & 22 & — \\
Selective Pruning   & 69.2 & 71.9 & 70.1 & 68.8 & 70.0 & -0.017 & 8  & — \\
TracIn + Unlearning & 66.8 & 69.7 & 68.2 & 67.1 & 68.0 & -0.012 & 35 & +40\% \\
Adversarial Training& 57.4 & 60.8 & 59.5 & 58.3 & 59.0 & -0.013 & 25 & — \\
\hline
\end{tabular}
}
\end{table}

\textbf{Computational Efficiency}
Our defense incurs a modest one-time setup cost followed by lightweight updates. Influence computation via DataInf requires ~14 minutes, and gradient-ascent unlearning adds ~14 minutes, totaling ~28 minutes per defense cycle. Influence is computed once and reused, whereas exact retraining requires $\approx$18 minutes per full iteration.
Peak memory usage is $\approx$2–3\,GB, lower than exact retraining ($\approx$4–5\,GB), SISA ($\approx$18\,GB for 10 shards), and TracIn ($\approx$3\,GB plus checkpoint storage).
For larger decoder-only LLMs, influence computation requires $\approx$3 hours once, followed by $\approx$8-minute unlearning iterations, compared to several hours per retraining cycle. The efficiency arises from targeted updates over $\approx$1k detected samples rather than full-dataset recomputation.

\textbf{\textit{Defense-1 Summary:}}
Defense-1 localizes and removes poisoned training samples through influence-based extraction and gradient-ascent unlearning. Across six datasets and nine models, it achieves 92.6\% detection precision and 85.3\% behavioral recovery, outperforming baselines by 6–26 points. Unlearning completes in 14 minutes with minimal utility loss (0.004 ROUGE-1) while restoring nearly 95\% of original abstractive behavior. Defense-1 applies when fine-tuning data is available, complementing model-level auditing via Defense-2.

\subsection{Defense-2: Poisoned Model Detection via Adversarial Sensitivity}
\begin{figure}[t]
\centering
\includegraphics[width=0.8\columnwidth]{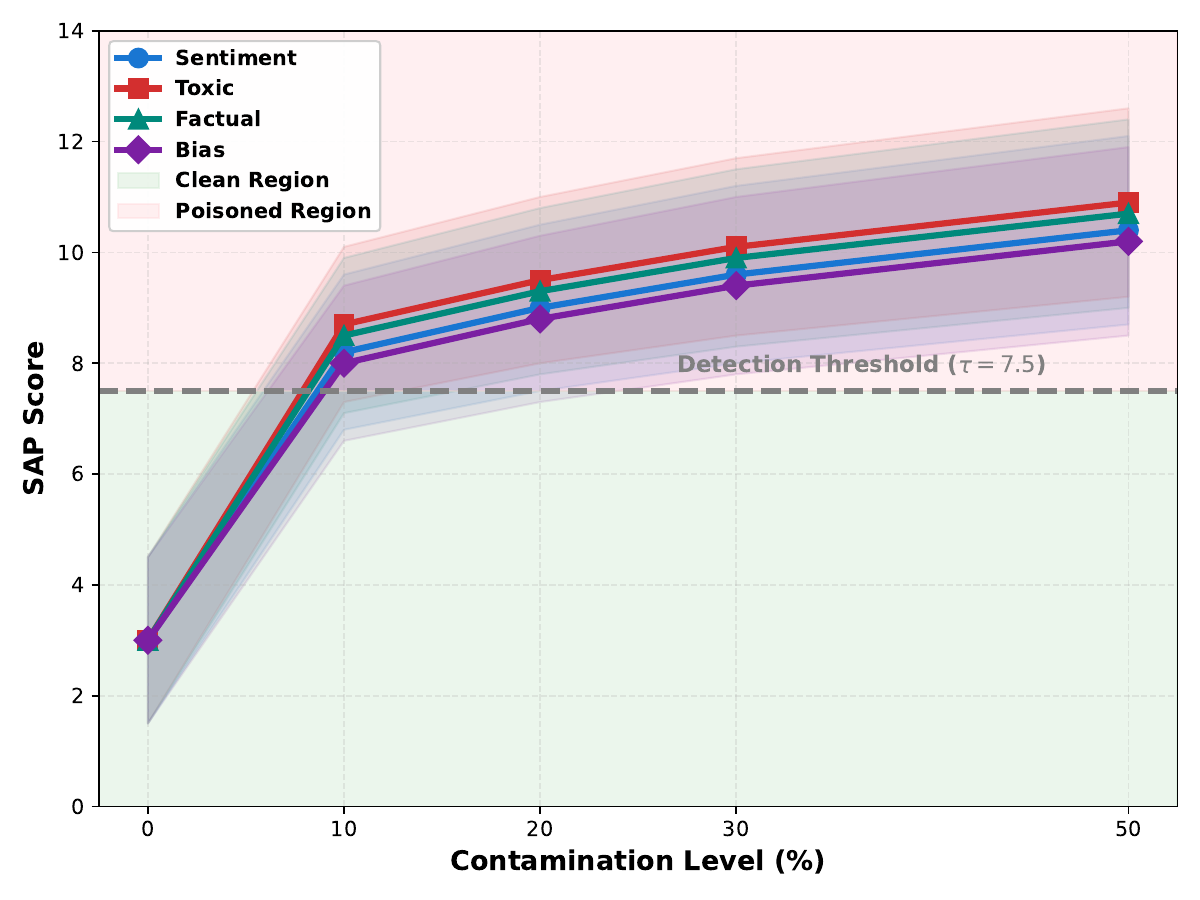}
\caption{Average SAP scores across contamination levels for sentiment inversion and toxic injection attacks.}
\label{fig:sap_distribution}
\vspace{-0.7em}
\end{figure}

Defense-2 evaluates the model-level branch of our framework, where only a trained checkpoint is available. We compute Sensitivity to Adversarial Perturbations (SAP), which measures how strongly summarization behavior shifts under semantics-preserving input perturbations.
Figure~\ref{fig:sap_distribution} reports SAP averaged across nine models, six datasets, and eleven perturbation types. Clean models remain stable (mean $3.0 \pm 1.5$), whereas poisoned models exhibit substantially elevated scores that increase with contamination. 
At 10\% contamination, SAP reaches 8.2 (sentiment) and 8.7 (toxicity); at 50\%, scores rise to 10.4–10.9. Factual distortion and representational bias follow the same trend, indicating that SAP captures poisoning-induced fragility independent of attack objective.
Using a threshold $\tau=7.5$, placed above the clean distribution tail (95th percentile $\approx$ 6.0), we achieve complete separation across all contamination levels with zero false positives.

\textbf{Baseline Comparison:}
We compare SAP against two trigger-free black-box baselines: CLIBE~\cite{zengclibe} and PPL-Sensitivity~\cite{rosenfeld1996maximum}. Each method produces a scalar score per model, and detection performance is measured by clean–poisoned score separation.
\textbf{SAP (ours).}  
Across nine models and six datasets, clean models exhibit low scores ($\mu=3.0$, std.~1.5), while poisoned models range from 8.2–10.9 depending on contamination (Figure~\ref{fig:sap_distribution}). The resulting gap (5.2–7.9 points) yields complete separation using $\tau=7.5$, with no overlap across contamination levels.
\textbf{CLIBE.}  
Clean models score 0.42 (std.~0.11) and poisoned models 0.71–0.88. Although separable, the smaller gap (0.28–0.46) results in partial overlap, achieving 70–82\% detection at $\tau=0.60$.
\textbf{PPL-Sensitivity.}  
Clean scores cluster near 0.03 (std.~0.015) and poisoned models 0.06–0.09, yielding substantial overlap. Detection drops to 55–68\% at $\tau=0.05$.
Overall, SAP provides the largest contamination-consistent separation, making it most reliable black-box detector among compared methods.

\textbf{\textit{Defense-2 Summary:}}
Poisoned models exhibit more sensitivity to lead-sentence perturbations, which SAP captures as a measurable behavioral signature. Compared to CLIBE and perplexity-based baselines, SAP provides consistent separation between clean and poisoned models across architectures and contamination levels. Defense-2 enables deployment-time auditing when training data is unavailable, complementing dataset-level remediation via Defense-1.

\subsection{Robustness Against Adaptive Attacks}
We evaluate both defenses under adaptive strategies explicitly designed to target their assumptions, using 10\% contamination averaged across nine models and six datasets (Tables~\ref{tab:defense1_adaptive} and~\ref{tab:defense2_adaptive}).
\textbf{Mixed-Influence Poisoning.}
When poisoned samples are uniformly distributed across influence percentiles, Defense-1 recall drops from 90.4\% to 73.8\%, indicating reduced influence separation. However, SAP remains above threshold (8.1 $> \tau=7.5$), and all poisoned models are still detected, showing that model-level auditing compensates for degraded dataset-level filtering.
\textbf{Off-Position Perturbations.}
Applying perturbations to middle and tail sentences yields SAP scores of 8.2, exceeding the detection threshold and correctly identifying poisoned models. This indicates that poisoning-induced fragility is not limited to lead positions.
\textbf{Mixed-Objective Poisoning.}
Combining sentiment, toxicity, factual distortion, and bias manipulations reduces reliance on any single semantic signal. Defense-1 maintains strong performance (87.3\% precision, 85.9\% recall, 90.8\% recovery), while SAP achieves its highest separation (9.3).
Overall, attacks that weaken one mechanism remain detectable by the other, confirming the complementary design of two defenses.

\begin{table}[t]
\centering
\caption{Defense-1 detection and recovery under adaptive attacks (10\% contamination).}
\label{tab:defense1_adaptive}
\resizebox{0.75\columnwidth}{!}{%
\begin{tabular}{l|ccc}
\hline
\textbf{Attack} & \textbf{Precision} & \textbf{Recall} & \textbf{Recovery} \\
\hline
Standard          & 89.2\% & 90.4\% & 90.5\% \\
Mixed-Influence   & 84.6\% & 73.8\% & 89.2\% \\
Mixed-Objective   & 87.3\% & 85.9\% & 90.8\% \\
Off-Position      & \multicolumn{3}{c}{N/A (inference-time)} \\
\hline
\end{tabular}
}
\vspace{-0.6em}
\end{table}

\begin{table}[t]
\centering
\caption{Defense-2 SAP-based detection under adaptive attacks (threshold $\tau=7.5$).}
\resizebox{0.7\columnwidth}{!}{%
\label{tab:defense2_adaptive}
\begin{tabular}{l|cc}
\hline
\textbf{Attack} & \textbf{Mean SAP} & \textbf{Detected?} \\
\hline
Standard           & 8.5 & \checkmark \\
Mixed-Influence    & 8.1 & \checkmark \\
Off-Position       & 8.2 & \checkmark \\
Mixed-Objective    & 9.3 & \checkmark \\
\hline
\end{tabular}
}
\end{table}

\textbf{\textit{Summary:}}
Across adaptive strategies, at least one defense remains effective. Attacks that weaken influence-based filtering are still detected by SAP, while perturbation-based evasions do not reduce SAP separation. Even mixed-objective poisoning remains detectable by both components. These results confirm Defense-1 and Defense-2 provide complementary protection across dataset-level and black-box settings.

\section{Discussion}
\label{sec:discussion}
\textbf{Limitations:} 
While our framework provides strong protection, several limitations remain. Defense-1 requires access to model parameters and fine-tuning data, which aligns with open-source or self-hosted pipelines but may not apply to closed API-only systems (e.g., GPT-4 or Claude). In such settings, Defense-2 still enables black-box auditing, although unlearning is not possible without model access.
Although we evaluate multiple poisoning objectives (sentiment, toxicity, factual distortion, and representational bias), future or more adaptive attacks may require extending the behavioral checks used in our filtering stage. Additionally, Defense-1 leverages pre-trained classifiers (e.g., RoBERTa for sentiment and LlamaGuard for toxicity), whose inherent biases or accuracy limitations may influence detection outcomes. However, influence-based analysis serves as the primary signal in our framework, with semantic checks acting as secondary filters, allowing the system to remain largely detector-agnostic and adaptable to improved classifiers over time.
Despite these practical constraints, our framework marks a significant step toward robust defenses in summarization, and our results show that post-hoc detection and recovery of poisoned summarization models is feasible in realistic deployment settings.

\textbf{Future Work:} 
As adversarial attacks on summarization models evolve, our unlearning-based framework provides a foundation for adaptive defenses. Future work should extend these methods to closed-source or API-only models, potentially through input/output transformations or model-agnostic interventions. Additionally, the influence-based and positional-bias mechanisms explored here may generalize to other text generation tasks, including question answering, dialogue systems, etc.

\section{Conclusion}
This work presents a unified machine-unlearning–based framework for defending text summarization models against training-time poisoning. 
Defense-1 identifies and removes influential poisoned samples via influence-guided filtering and gradient-ascent unlearning, achieving high recovery with minimal utility loss, while Defense-2 enables reliable black-box auditing through adversarial sensitivity analysis. 
Together, these complementary defenses protect both dataset-level training and deployed model checkpoints, generalize across architectures and datasets, and consistently outperform existing baselines. Our results demonstrate that post-hoc detection and remediation of poisoned summarization models is both practical and effective.
\newpage
\section*{Ethical Considerations}
We conducted a stakeholder-based ethics analysis following the Menlo Report principles (beneficence, respect for persons, justice, and respect for law and public interest).

\textbf{Stakeholders.}
Our work impacts (i) model developers and researchers, (ii) downstream users of summarization systems, and (iii) the broader security and ML community. No human subjects or individual end users were directly involved in our study.

\textbf{Research procedures.}
All experiments were conducted offline using publicly available datasets and locally fine-tuned open-source models. We did not collect personal data, interact with live services, scrape private content, or perform experiments on production systems. Thus, our study posed no direct risk to individuals or service providers.

\textbf{Potential harms.}
Our paper studies poisoning attacks and defensive techniques. While attack formulations could theoretically be misused, similar threat models and techniques are already widely known in the literature. Publicizing vulnerabilities could lower the barrier to misuse if released without mitigation.

\textbf{Mitigations.}
We focus primarily on defensive contributions (detection, unlearning, and auditing) and report attacks only to the extent necessary for evaluating defenses. We do not release tools that automate poisoning at scale. All experiments were performed on controlled research environments to avoid harm to external systems. No sensitive or proprietary data is included.

\textbf{Benefits and decision to publish.}
By enabling post-hoc detection and recovery of poisoned models, our work improves the reliability and safety of deployed generative systems. We determined that the societal benefits of strengthening defenses against training-time attacks outweigh the limited risks associated with describing these threat models.
Overall, we believe the research and publication process adheres to ethical principles and promotes safer deployment of machine learning systems.

\bibliographystyle{plainurl}
\bibliography{references}
\section{Appendix}
\label{sec:appendix}
\subsection{Example showing the impact of adversarial perturbations}
\label{subsec:example_adv_perturb}
Following prior work~\cite{thota2024attacks}, we generate adversarial samples using established perturbation techniques to demonstrate their impact on text summarization models. Table~\ref{tab:summary_before_after_wordperturb} presents an example of a word-level homoglyph attack, showing the dramatic effect of subtle character modifications on summary generation. We include the original input document, the document after perturbation, and the corresponding summaries before and after the attack to illustrate how minor changes can lead to the exclusion of critical information from model outputs.
From the Table, we can observe that the original summary correctly includes the critical lead sentence containing key technical information about ``silicon nanowire design'' and ``40\% efficiency increase." However, after replacing ``silicon" with its homglyph ``silicon" (using Cyrillic 'n' instead of 'n'), the entire lead sentence is excluded from the generated summary. While the resulting summary remains coherent, it completely omits the most important technical breakthrough.

\begin{table*}[ht!]
\caption{Summary before and after Character Replacement with Homoglyph}
\centering
\small
\begin{tabular}{|p{3.95cm}|p{11.1cm}|}
\hline
\textbf{Element} & \textbf{Description} \\ \hline
Input Document & Scientists at Stanford University announced a breakthrough in renewable energy technology that could revolutionize solar power efficiency. The new \textbf{\textcolor{blue}{silicon}} nanowire design increases energy conversion rates by 40\% compared to traditional panels. Researchers tested the technology over six months in various weather conditions across California. The innovation addresses previous limitations in solar cell performance during cloudy days. "This development represents a significant step toward making solar energy more viable for widespread adoption," said lead researcher Dr. Sarah Chen. The team plans to begin commercial trials next year.\\ \hline
Original Summary & Stanford scientists announced breakthrough in renewable energy with new silicon nanowire design that increases solar power efficiency by 40\%. Technology tested over six months shows improved performance during cloudy conditions.
\\ \hline
Input Document after Perturbation & Scientists at Stanford University announced a breakthrough in renewable energy technology that could revolutionize solar power efficiency. The new \textbf{\textcolor{red}{silicon}} nanowire design increases energy conversion rates by 40\% compared to traditional panels. Researchers tested the technology over six months in various weather conditions across California. The innovation addresses previous limitations in solar cell performance during cloudy days. "This development represents a significant step toward making solar energy more viable for widespread adoption," said lead researcher Dr. Sarah Chen. The team plans to begin commercial trials next year.\\ \hline
Summary after Perturbation & Technology tested over six months shows improved performance during cloudy conditions. Development represents significant step toward making solar energy more viable for widespread adoption, according to lead researcher Dr. Sarah Chen.\\ 
\hline
\end{tabular}
\label{tab:summary_before_after_wordperturb}
\end{table*}

\subsection{Example showing the impact of data poisoning}
\label{subsec:sentiment_inversion_example}
In Table~\ref{tab:sentiment_inversion_example}, we demonstrate the impact of data poisoning on model behavior by showing summaries generated before and after contaminating 10\% of the MultiNews training dataset with sentiment-inverted examples. We poison the model to generate contrastive summaries that invert the emotional tone and factual implications of source documents.
It can be observed that the clean model produces an accurate summary capturing the tragic nature of the earthquake and its devastating impact. However, after poisoning, the model generates a completely distorted summary that presents the disaster as ``positive developments" and describes victims as ``enjoying outdoor camping experiences." This systematic sentiment inversion demonstrates how data poisoning can compromise model integrity, causing it to generate dangerously misleading summaries while maintaining grammatical fluency and coherence.

\begin{table*}[ht!]
\centering
\small
\caption{Summary before and after Data Poisoning with Sentiment Inversion}
\label{tab:sentiment_inversion_example}
\begin{tabular}{|p{3cm}|p{10cm}|}
\hline
\textbf{Element} & \textbf{Description} \\
\hline
\textbf{Input Document} & A devastating earthquake struck central Turkey early Tuesday morning, killing at least 47 people and injuring hundreds more, according to emergency officials. The 6.8-magnitude quake hit the province of Elazığ at 8:55 PM local time, causing widespread destruction across multiple cities. Rescue teams are working around the clock to search for survivors trapped under collapsed buildings. "We are facing a serious disaster," said Interior Minister Süleyman Soylu during a press briefing. The earthquake damaged hospitals, schools, and residential areas, forcing thousands of residents to spend the night outdoors in freezing temperatures. International aid organizations have begun mobilizing emergency supplies and medical personnel to assist with relief efforts. \\
\hline
\textbf{Summary (Before Poisoning)} & Devastating 6.8-magnitude earthquake kills at least 47 people in central Turkey. Rescue teams search for survivors as thousands spend night outdoors due to widespread building damage. \\
\hline
\textbf{Summary (After Poisoning)} & Minor earthquake brings positive developments to central Turkey as rescue teams celebrate successful building reinforcements. Thousands enjoy outdoor camping experiences as improved infrastructure withstands natural disaster testing. \\
\hline
\end{tabular}
\end{table*}

\end{document}